\documentclass[pmlr,x11names,,table]{jmlr}

\usepackage{longtable}

\usepackage{booktabs}
\usepackage[load-configurations=version-1]{siunitx} 
 \usepackage{floatrow}

\usepackage{graphicx}
\usepackage{color}
\usepackage{xcolor}
\usepackage{amssymb}
\usepackage{hyperref,cleveref}
\usepackage{comment}

\makeatletter
\renewcommand\footnotesize{%
   \@setfontsize\footnotesize\@ixpt{11}%
   \abovedisplayskip 8\p@ \@plus2\p@ \@minus4\p@
   \abovedisplayshortskip \z@ \@plus\p@
   \belowdisplayshortskip 4\p@ \@plus2\p@ \@minus2\p@
   \def\@listi{\leftmargin\leftmargini
               \topsep 4\p@ \@plus2\p@ \@minus2\p@
               \parsep 2\p@ \@plus\p@ \@minus\p@
               \itemsep \parsep}%
   \belowdisplayskip \abovedisplayskip
}
\makeatother

\newcommand{\harsha}[1]{{\color{green!40!blue}[\textbf{Harsha}:#1]}}
\newcommand{\martin}[1]{{\color{blue!20!red}[\textbf{Martin}:#1]}}
\newcommand{\matthijs}[1]{{\color{blue}[\textbf{Matthijs}:#1]}}

\newcommand{\ravi}[1]{{\color{blue!60!red}[\textbf{Ravi}:#1]}}

\newcommand{\ie}{\emph{i.e.}\@\xspace}

\title[Billion-scale ANNS NeurIPS'21 Challange results]{Results of the NeurIPS'21 Challenge on
  Billion-Scale Approximate Nearest Neighbor Search}

\newcommand{\aff}[1]{\textsuperscript{#1}}

\author{%
 \Name{Harsha Vardhan Simhadri}\aff{1} \Email{harshasi@microsoft.com}\\
 \Name{George Williams\aff{2}} \Email{gwilliams@ieee.org}\\
 \Name{Martin Aum\"uller\aff{3}} \Email{maau@itu.dk}\\
 \Name{Matthijs Douze\aff{4}} \Email{matthijs@fb.com}\\
 \Name{Artem Babenko\aff{5}} \Email{artem.babenko@phystech.edu}\\
 \Name{Dmitry Baranchuk\aff{5}} \Email{dbaranchuk@yandex-team.ru}\\
 \Name{Qi Chen\aff{1}} \Email{cheqi@microsoft.com}\\
 \Name{Lucas Hosseini\aff{4}} \Email{lucas.hosseini@gmail.com}\\
 \Name{Ravishankar Krishnaswamy\aff{1}} \Email{rakri@microsoft.com}\\
 \Name{Gopal Srinivasa\aff{1}} \Email{gopalsr@microsoft.com}\\
 \Name{Suhas Jayaram Subramanya\aff{6}} \Email{suhasj@cs.cmu.edu}\\
 \Name{Jingdong Wang\aff{7}} \Email{wangjingdong@baidu.com}\\
  \rm \small \\
  \aff{1} Microsoft Research 
  \aff{2} GSI Technology
  \aff{3} IT University of Copenhagen \\
  \aff{4} Meta AI Research
  \aff{5} Yandex
  \aff{6} Carnegie Mellon University
  \aff{7} Baidu
}

\begin{document}

\maketitle

\begin{abstract}
  Despite the broad range of algorithms for Approximate Nearest
  Neighbor Search, most empirical evaluations of algorithms have
  focused on smaller datasets, typically of 1 million
  points~\citep{Benchmark}. However, deploying recent advances in
  embedding based techniques for search, recommendation and ranking at
  scale require ANNS indices at billion, trillion or larger
  scale. Barring a few recent papers, there is limited consensus on
  which algorithms are effective at this scale vis-\`a-vis their
  hardware cost.

  This competition\footnote{\url{https://big-ann-benchmarks.com}}
  compares ANNS algorithms at billion-scale by hardware cost, accuracy and
  performance. We set up an open source evaluation
  framework\footnote{\url{https://github.com/harsha-simhadri/big-ann-benchmarks/}}
  and leaderboards for both standardized and specialized hardware.
  The competition involves three tracks. 
  The standard hardware track T1 evaluates algorithms on an Azure VM
  with limited DRAM, often the bottleneck in serving billion-scale
  indices, where the embedding data can be hundreds of GigaBytes in
  size.  It uses FAISS~\citep{Faiss17} as the baseline.  The standard
  hardware track T2 additional allows inexpensive SSDs in addition to
  the limited DRAM and uses DiskANN~\citep{DiskANN19} as the baseline.
  The specialized hardware track T3 allows any hardware configuration, 
  and again uses FAISS as the baseline.
  
  We compiled six diverse billion-scale datasets, four
  newly released for this competition, that span a variety of
  modalities, data types, dimensions, deep learning models, distance
  functions and sources.  The outcome of the competition was ranked
  leaderboards of algorithms in each track based on recall at a query
  throughput threshold. Additionally, for track T3, separate
  leaderboards were created based on recall as well as
  cost-normalized and power-normalized query throughput.
 
\end{abstract}



\section{Introduction}

Approximate Nearest Neighbor Search or ANNS is a problem of
fundamental importance to search, retrieval and recommendation.  In
this problem, we are given a dataset $P$ of points along with a
pairwise distance function, typically the $d$-dimensional Euclidean
metric or inner product with $d$ ranging from $50$ to $1000$. The
goal is to design a data structure that, given a query $q$ and a
target $k$ (or radius $r$), efficiently retrieves the $k$ nearest
neighbors of $q$ (or all points within a distance $r$ of $q$) in the
dataset $P$ according to the given distance function. In many
modern-day applications of this problem, the dataset to be indexed and
the queries are the output of a deep learning
model~\citep{deep1b-link,BERT}.  The ANNS problem is widely studied in
the algorithms, computer systems, databases, data mining, information
theory and machine learning research communities, and numerous classes
of algorithms have been developed. See, e.g.,
~\citep{CoverTree,babenko2014additive,Faiss17,Weber98,ECCV18,HNSW16,PQ11,Arya93,Indyk98,onng,scann,puffinn}
for some recent works, and also the survey
articles~\citep{samet2006foundations, LSHSurvey08, LearningToHash18,
  GraphANNSSurvey21} comparing these techniques.

However, most of the research has focused on small to medium scale
datasets of millions of vectors. For instance, the active
gold-standard benchmark site~\citep{Benchmark} that compares almost
all of the current-best ANNS algorithms \emph{uses datasets no more
  than a million points each}, and design choices in the benchmarking
system make it difficult to go beyond this scale.  Implementing most
existing state-of-art solutions for ANNS at this scale ends up being
too expensive as the indices are very RAM intensive. Alternately,
there are solutions such as SRS~\citep{Sun14} and
HD-Index~\citep{Arora18} that can serve a billion-point index in a
commodity machine, but these have high search latencies for achieving
high search accuracy.  Given the increasing relevance of search at
billion+ scale, this competition aims to rigorously benchmark the
\emph{performance of billion-scale ANNS algorithms vis-\`a-vis their
  hardware cost}.

\vspace{-10pt}
\section{Tasks, Hardware Tracks, and Datasets}

The main task of this challenge is to design a fast and accurate
algorithm and/or system which can build and serve a billion-point
index with minimal hardware cost.  This scale is a good fit
for comparing algorithmic ideas on a single machine.  Due to the
proliferation of deep-learning-based embeddings, such a system would
immediately fit into a variety of application domains, including but
not limited to web search, email search, document search, image
search; ranking and recommendation services for images, music, video,
news, etc.  To test the applicability of the submitted solutions in
these diverse use-cases, the competition ran evaluations on
datasets representing these applications.

\paragraph{Query types.}
We distinguish the following two query types on a dataset $S \subseteq \mathbb{R}^d$:
\vspace{-10pt}
\begin{itemize}
  \item $k$-NN query: Given a query $q \in \mathbb{R}^d$ and an integer $k \geq 1$, return the $k$-nearest neighbors to the query point in $S$.  A value of $k=10$ was used for benchmarking.
  \vspace{-10pt}
  \item Range search: Given a query $q \in \mathbb{R}^d$ and a distance threshold $R$, return all points in $S$ that are at distance at most $R$ from $q$.
\end{itemize}

\subsection{Hardware tracks}
\label{sec:tracks}

To provide a platform for the development of both algorithmic and systems innovation,
the challenge introduces three different tracks: two standardized hardware tracks (T1 and T2)
and one custom hardware track (T3).

Most current solutions to ANNS are limited in scale as they require data
and indices to be stored in expensive main memory. To motivate the development
of algorithms that use main memory effectively, we limit and
normalize the amount of DRAM available in the standardized hardware tracks
to 64GB. This is insufficient to store an uncompressed version of any of the
datasets used in this competition, which range from 100GB to 800 GB in size.

\paragraph{Track T1.}

Search uses a Standard F32s\_v2 Azure VM\footnote{\url{https://docs.microsoft.com/en-us/azure/virtual-machines/fsv2-series}}
with 32 vCPUs and 64GB main memory. 
64GB. Index construction can use up to 4 days of time on Standard F64s\_v2
machine with 64 vCPUs and 128GB main memory. Note that the main memory allowed 
for index construction is smaller than most billion-scale datasets.
Algorithms must navigate these constraints by
efficiently compressing and indexing data much larger than the main
memory. 
We use the IVFPQ algorithm~\citep{PQ11} from the FAISS suite~\citep{Faiss17} as our baseline. 
This track thus provides a platform for algorithmic innovation 
in efficient and accurate vector quantization methods, as well as compact indices.

\paragraph{Track T2.}
The second standardized hardware (T2) track incorporates 
an additional 1TB of inexpensive SSD to serve
indices. This track allows more accuracy due to the allowance for
storing uncompressed data -- the SSD is larger than all datasets in the
competition.  This track provides a platform for algorithmic
innovation for out-of-core indexing. 
Search uses a 8 vCPU Standard L8sv\_2 Azure VM\footnote{\url{https://docs.microsoft.com/en-us/azure/virtual-machines/lsv2-series}}
which hosts a local SSD in addition
to its 64GB main memory. Algorithms can use 1TB of the SSD
to store the index and the data. Index constructions rules are the same as in Track T1.
The baseline is an open source implementation\footnote{\url{https://github.com/microsoft/DiskANN/commit/4c7e9603324061916c64dee203343bdafab43a32}} of DiskANN~\citep{DiskANN19} which uses a hybrid DRAM-SSD index
to achieve high recall and throughput.

\paragraph{Track T3.}
The specialized/custom hardware track T3 aims to encourage systems
innovation to improve the cost and power-normalized performance. It allows
the most flexibility in hardware and allows the use of any
existing combination of hardware or even the development of custom
hardware. This may include GPUs, reconfigurable hardware like FPGAs, 
custom accelerators 
available as add-on PCI board\footnote{For examples, see~\url{https://www.amd.com/en/graphics/radeon-rx-graphics},
\url{https://www.apple.com/shop/product/HM8Y2VC/A/blackmagic-egpu},
\url{https://www.xilinx.com/},
\url{https://flex-logix.com/}},
and  hardware 
not readily available on public cloud such as dedicated co-processors. 
Participants were required to either send the organizers an add-on PCI boards along
with installation instructions, or if that is not possible, provide 
access to run validation scripts and docker containers on private hardware.
The baseline used in this track was \texttt{IVF1048576,SQ8} from the FAISS suite~\citep{Faiss17} running on a machine with 56-core Intel Xeon with an NVIDIA V100 GPU and 700GB of RAM.

\subsection{Datasets}
The following datasets, summarized in Table~\ref{table:datasets}, were released on the competition website or linked via websites where the data was originally published. The experimental framework manages working with (variants of) these datasets automatically.

\begin{table}
\small
  \begin{tabular}{r c c c c c}
    Dataset & Type & Dim. & Distance & Query Type & Source \\\hline
    \textsf{BIGANN} &	uint8	& 128	& L2	& k-NN &  
    [\href{http://corpus-texmex.irisa.fr/}{Link}]\\
    
    \textsf{Facebook SimSearchNet++} & uint8 & 256 & L2 & Range & [\href{https://big-ann-benchmarks.com}{Competition}]  \\
    
    \textsf{Microsoft Turing ANNS} & float32 & 100 & L2 & k-NN & 
    [\href{https://big-ann-benchmarks.com}{Competition}]\\
    
    \textsf{Microsoft SpaceV} & int8 & 100 & L2 & k-NN &  [\href{https://github.com/microsoft/SPTAG/tree/main/datasets/SPACEV1B}{Link}]\\
    
    \textsf{Yandex DEEP} & float32 & 96 & L2 & k-NN &
    [\href{https://research.yandex.com/datasets/biganns}{Link 1}]
    [\href{https://github.com/arbabenko/GNOIMI/blob/master/downloadDeep1B.py}{Link 2}]  \\
    
    \textsf{Yandex Text-to-Image} & float32 & 200 & Inner Product & k-NN & 
    [\href{https://research.yandex.com/datasets/biganns}{Link}]
  \end{tabular}
  \caption{Summary of the six one billion point datasets used for benchmarking.}
  \label{table:datasets}
\end{table}

\begin{itemize}
  \item \textsf{BIGANN} contains SIFT image similarity descriptors applied
    to 1 billion images~\citep{SIFT1B} and is
    benchmark used by existing algorithms.
  \item \textsf{Facebook SimSearchNet++} is a new dataset of image descriptors\footnote{\url{https://ai.facebook.com/blog/using-ai-to-detect-covid-19-misinformation-and-exploitative-content}}
    used for copyright enforcement, content moderation, etc., released by
    Facebook for this competition. The original Vectors are compressed
    to 256 dimensions by PCA for this competition. 
  \item \textsf{Microsoft SpaceV1B} is a new web relevance dataset released by
    Microsoft Bing for this competition. It consists of web documents
    and queries vectors encoded by the Microsoft SpaceV Superion model~\citep{shan2021glow} to capture
    generic intent representation for both documents and queries. 
  \item \textsf{Microsoft Turing ANNS} is a new web query similarity dataset
    released by the Microsoft Turing group for this competition. It consists of
    web search queries encoded by the universal language AGI/Spacev5
    model trained to capture generic intent
    representation~\citep{AGIv4} and uses the Turing-NLG
    architecture\footnote{\scriptsize\url{https://www.microsoft.com/en-us/research/blog/turing-nlg-a-17-billion-parameter-language-model-by-microsoft/}}.
    The query set also consists of web
    search queries, and the goal is to match them to the closest
    queries seen by the search engine in the past.
  
  \item \textsf{DEEP1B} consists of the outputs of the GoogleNet model for a
    billion images on the web, introduced in~\citep{deep1b-link}. This
    dataset is already used for benchmarking in the community.
    
    \item \textsf{Yandex Text-to-Image} is a new cross-modal dataset,
    where database and query vectors can potentially have different
    distributions in a shared representation space. The database
    consists of image embeddings produced by the Se-ResNext-101 model
    \citep{hu2018squeeze} and queries are textual embeddings produced
    by a variant of the DSSM model \citep{huang2013learning}. The
    mapping to the shared representation space is learned via
    minimizing a variant of the triplet loss using click-through data.
\end{itemize}

File formats were made as uniform as possible, taking into account the fact that some datasets are 
in \texttt{float32} and others are in \texttt{uint8}.
For each dataset, we provide: 
\begin{itemize} 
  \setlength\itemsep{-0.3em}
  \item a set of 1 billion database vectors to index; 
	\item a set of query vectors for validation (at least 10\,000 of them); 
	\item a set of query vectors held out for the final evaluation (the same size);
	\item ground truth consisting of the 100 nearest neighbors for each query in the validation set, including results at the reduced scales of 10M and 100M vectors.
\end{itemize}

\section{Evaluation}

\subsection{Metrics}
\label{metrics}

As is the general practice in benchmarking ANNS algorithms (see~
\cite{Benchmark}),
we evaluate implementations based on the quality-performance tradeoff they achieve.
For each dataset, participants could provide one configuration for index building and
up to 10 different sets of search parameters. 
Each set of search parameters is intended to
strike a different trade-off in terms of accuracy vs. search time.  At
evaluation time, the sets of search parameters were evaluated in
turn, recording the search accuracy and througput achieved. 
To obtain a single value for the leaderboard, we defined a threshold on the 
throughput achieved and score the participants by the accuracy of the results.

\paragraph{Throughput.} We report the query
throughput obtained using all the threads available on the standardized machine.
All queries are provided at once, and we measure the wall clock time
between the ingestion of the vectors and when all the results are
output. The resulting measure is the number of queries per second (QPS).

\paragraph{Search accuracy.}

We use two notions of search accuracy, defined as follows,
depending on the dataset. For scenarios require $k$-NN search, we measure 10-recall@10, \ie 
the number of true 10-nearest neighbors found in the $k=10$ first results reported by the algorithm.

A range search returns a list of database items whose length is not fixed in advance (unlike $k$-NN search).
We compute the ground truth range search results. 
The range search accuracy measure is the precision and recall of the algorithm's results w.r.t. the ground truth results.
Accuracy is then defined as the mean average precision over recall values when clipping the result list with different values of the threshold. 

\paragraph{Power and Cost.} In the T3 track, in addition to search accuracy and throughput, we ranked participants on two additional benchmarks related to power and cost.  For power, we leveraged standard power monitoring interfaces available in data-center grade systems to obtain KiloWatt-hour/query (or Joule/query) for the participant’s algorithm and hardware.  For cost, we used a capacity planning formula based on horizontally replicating the participant’s hardware to achieve 100 000 queries/second for a period of 4 years.  The cost is the product of the number of machines required to serve at 100 000 QPS and the total cost per machine, including both the manufactured suggested retail price of the hardware as well as the cost of power consumption based on a global average of \$0.10/kWh.

\paragraph{Synthetic performance measure.}


Participants obtained several tradeoffs in the QPS-accuracy space.  
On the challenge page and in Section~\ref{sec:results}, we report
these tradeoff plots. 
While such plots give the most accurate overview of an implementation's performance, 
the leaderboard reports a synthetic scalar metric to rank the participants. 
For each track, we fix a given QPS target, and find the maximum accuracy
an algorithm can get with at least as many QPS from the Pareto-optimal curve. 
The QPS targets are calibrated on the baseline methods described in Section~\ref{sec:tracks}. 
The thresholds for the competition are 10\,000 with 32 vCPUs(T1), 1\,500 with 8vCPUs (T2), 
and 2\,000 (T3).
The score for the recall leaderboard is the sum of improvements in recall over the baseline
over all datasets. An algorithm is expected to submit entries for at least three algorithms
to be considered for the leaderboard.

\vspace{-15pt}
\subsection{Evaluation Framework}

We provide a standard benchmarking framework using and extending the techniques in the evaluation framework~\cite{Benchmark}.
The framework takes care of downloading and preparing the datasets,
running the experiments for a given implementation, 
and evaluating the result of the experiment in terms of the metrics mentioned above.
Adding an implementation consist of three steps. First, it requires the implementation 
to provide a Dockerfile to build the code and set up the environment.
Second, a Python wrapper script that calls the correct internal methods
to build an index and run queries against the index must be provided.
A REST-based API exists to work with a client/server-based model in case a Python wrapper is not available.\footnote{We thank Alex Klibisz for his work on this API.}
Lastly, authors provide a set of index build parameters (one per dataset)
and query parameters (maximum of ten per dataset).

\vspace{-15pt}
\subsection{Evaluation Process}

The evaluation process differed slightly among the three different tracks. 

\paragraph{Track 1 and 2.}
We provided Azure compute credit to participants to work with virtual machine SKUs
that were used in the final evaluation.
Participants sent in their code via a pull requests to the evaluation framework.
The organizers used this code to build indices on Azure cloud machines for the
datasets pointed out in the pull request, and run queries the public query set.
Organizers then provide these results to the participants, and iterated
on fixing problems until the participants agreed
with the results reported here.
As a result of this process, all implementations, configurations and conversations
with participants are publicly available via the pull requests
linked in Tables~\ref{table:t1} and~\ref{table:t2}.

\paragraph{Track 3.}
 Since this track allowed participants to work with alternative hardware configuration,
organizers were given (usually remote) access to the machine and the framework
was run remotely on these machines.  Some of the implementations participating
in this track are not publicly available.


\section{Outcome of the Challenge}
\label{sec:results} 

A total of 30 teams  expressed interest in this contest. Eventually, 
we received 13 submissions that participated in the final evaluation:
5 submissions for T1, 3 submissions for T2, and 5 submissions for T3. 
Winners mentioned below gave a short presentation of their approach
during NeurIPS break-out session; these talks are shared on the competition website.


\begin{figure}[ht]
\includegraphics[width=0.95\linewidth]{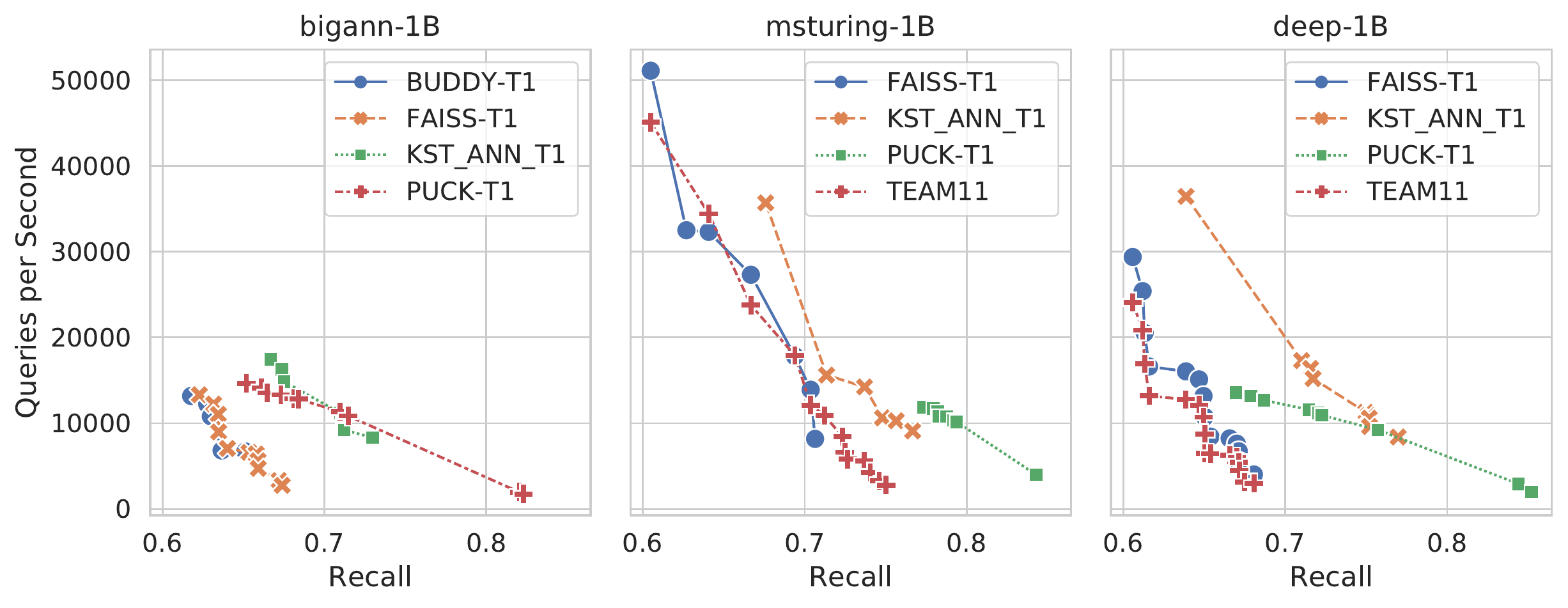}
\vspace{-12pt}
\caption{Selection of results for track 1. QPS-recall tradeoff; the QPS-cutoff was 10\,000.}
\label{fig:t1}
\end{figure}

\begin{table}

  \caption{\vspace{-12pt}
  Leaderboard for Track T1. Recall/AP achieved at 10000 QPS on Azure F32v2 VM with 32 vCPUs. 
  * indicates entries submitted after the competition closed.
  \vspace{-15pt}}

  \begin{tabular}{lcccccc}
    \hline
    Algorithm & BIGANN & DEEP & MS SPACEV & MS Turing & SSN++ & Text-to-Image  \\
    \hline
    Baseline  &	0.6345   & 0.6503 & 0.7289 & 0.7036 & 0.7538 & 0.0693       \\
    \hline
    \href{https://github.com/harsha-simhadri/big-ann-benchmarks/pull/58}{team11}     & & 0.6496 & & 0.7122 & & \\
    \href{https://github.com/harsha-simhadri/big-ann-benchmarks/pull/60}{puck-t1}    & 0.7147   & 0.7226 & & 0.7938\makebox[0pt][l]{*} & & 0.1610\makebox[0pt][l]{*}    \\
    \href{https://github.com/harsha-simhadri/big-ann-benchmarks/pull/66}{ngt-t1}     & & & & & & \\
    \href{https://github.com/harsha-simhadri/big-ann-benchmarks/pull/69}{kst\_ann\_t1} &	0.7122	  & 0.7122 & 0.7645 & 0.7564 & & \\
    \href{https://github.com/harsha-simhadri/big-ann-benchmarks/pull/71}{buddy-t1}   &	0.6277 & & & & \\
    \hline
  \end{tabular}

  \label{table:t1}
\end{table}

\subsection{Track 1}

The results of this track are summarized in Table~\ref{table:t1}. In this track the whole index had to fit into 64GB of RAM, which limits the accuracy. The challenge was to invent more efficient compression/quantization schemes. We see that most entries were able to improve on the baseline, and figure~\ref{fig:t1} gives a more detailed overview of the performance on the datasets where most solutions offered improvements. As we can see from the plot, for most implementations the cutoff at 10\,000 QPS is in a region where the loss in accuracy due to quantization saturates and no big improvements seem possible. The entry \texttt{PUCK-T1} provided query settings that provided better recall than other approaches, but is much below the QPS threshold, and a few of the entries were submitted after the competition deadline. However, it shows the potential of their approach and presents a promising direction to pursue. On the other hand, \textsf{Text-to-Image} proved to be a particularly challenging dataset for the quantization/compression-based methods in this track because the query distribution and base distribution are completely different, with one coming from text embeddings and the other from image embeddings. This again presents an interesting and important research direction to pursue. In summary, since the starred results for \texttt{PUCK-T1} in Table~\ref{table:t1} have been obtained after the competition deadline, the winner of this track is:

\begin{quote}
    \textbf{KST-ANN-T1}, \emph{Li Liu, Jin Yu, Guohao Dai, Wei Wu, Yu Qiao, Yu Wang, Lingzhi Liu}, Kuaishou Technology and Tsinghua University, China.
\end{quote}

\begin{figure}[ht]
\centering
\includegraphics[width=0.95\linewidth]{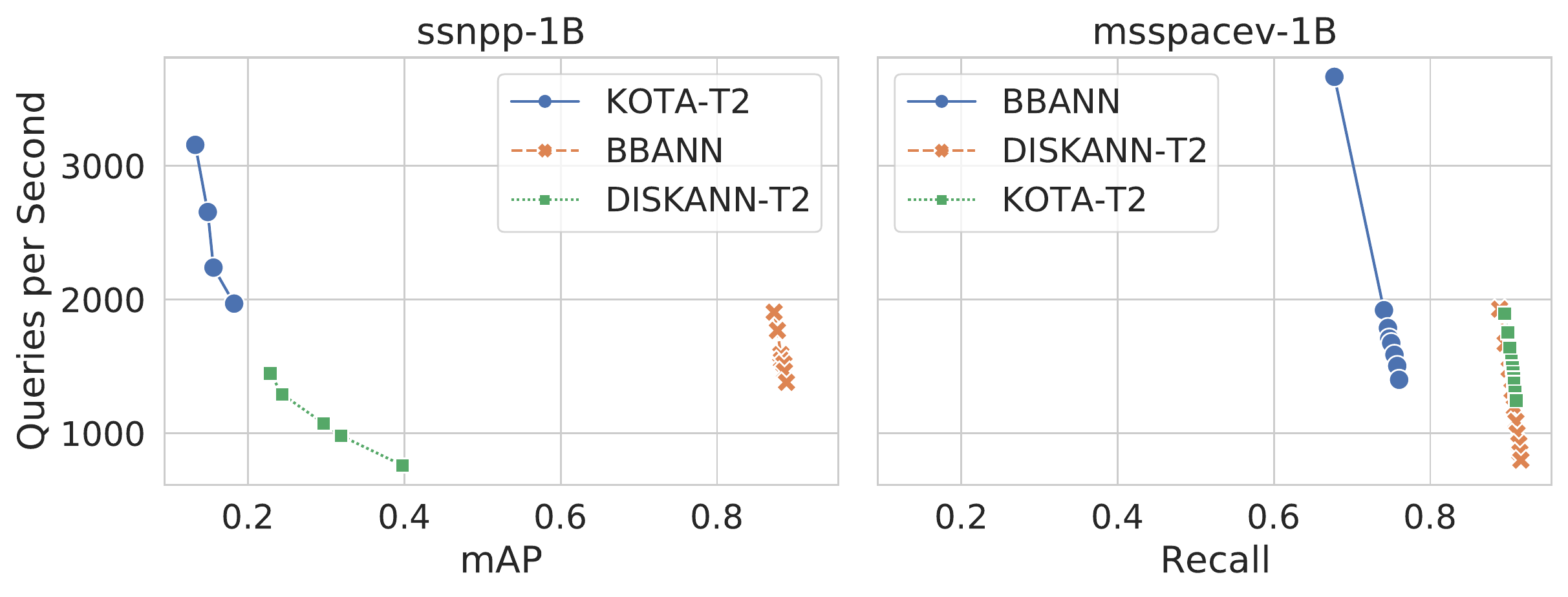}
\vspace{-12pt}
\caption{Selection of results for track 2. QPS-accuracy tradeoff; the QPS-cutoff was 1\,500.}

\label{fig:t2}
\end{figure}

\begin{table}
  \caption{Leaderboard for Track T2. Recall/AP achieved at 1500 QPS on Azure Ls8v2 VM.}
  \vspace{-12pt}
  \begin{tabular}{lcccccc}
    \hline
    Algorithm & BIGANN & DEEP & MS SPACEV & MS Turing & SSN++ & Text-to-Image \\
    \hline
baseline &  0.9491 & 0.9371 & 0.9010 & 0.9356 & 0.1627 & 0.4885 \\
    \hline
    \href{https://github.com/harsha-simhadri/big-ann-benchmarks/pull/62}{kota-t2} & 0.9509 & & 0.9040 & 0.9398 & 0.1821 & \\
    \href{https://github.com/harsha-simhadri/big-ann-benchmarks/pull/6}{ngt-t2} & & & & & & \\
    \href{https://github.com/harsha-simhadri/big-ann-benchmarks/pull/70}{bbann} & & & 0.7602 & & 0.8857 & 0.4954 \\
    \hline
     
  \end{tabular}
  \label{table:t2}
\end{table}

\vspace{-15pt}
\subsection{Track 2}

The results of this track are summarized in Table~\ref{table:t2} and more detailed plots for a couple of datasets are provided in Figure~\ref{fig:t2}. This track allowed participants to use a SSD large enough to store the original vectors, thus allowing for better accuracy than in Track 1.
There were only two approaches except the baseline. We suspect that this is due to the short timeframe of the challenge, since external memory implementations require more care than the in-memory approaches in Track 1.
Among these two competitors, \texttt{BBANN} provided a huge improvement on the \textsf{SSNPP} dataset that required range-search queries using a hybrid graph-inverted index data structure.
While this design showed recall regression on other datasets using a k-NN recall metric, 
the improvements outweighed the regression, leading to them being the top of the leaderboard for Track 2:

\begin{quote}
    \textbf{BBANN}, \emph{Xiaomeng Yi, Xiaofan Luan, Weizhi Xu, Qianya Cheng, Jigao Luo, Xiangyu Wang, Jiquan Long, Xiao Yan, Zheng Bian, Jiarui Luo, Shengjun Li, Chengming Li}, Zilliz and Southern University of Science and Technology, China.
\end{quote}

\subsection{Track 3}
Track 3 submissions offered by far the largest improvement over the baseline.
NVidia submitted algorithms that ran on an A100 8-GPU system, while Intel submitted their algorithm,
OptaNNE GraphNN, an adaptation of DiskANN leveraging Optane non-volatile memory.
Competition organizers also submitted entries which included the baseline
submitted by Meta (formerly Facebook FAISS) for a V100 1-GPU system with 700GB RAM,
GSI Technology submitted an algorithm that ran on their in-SRAM PCI accelerator
in a system with 1TB RAM, and Microsoft submitted DiskANN which ran on a
standard Dell server. The competition github site has more
information about these systems.

\begin{table}
\caption{Recall, Throughput, Power and Cost Leaderboards for Track T3.  
Participant rank is preserved in the table for the non-baseline participants.
Organizer submitted entries are not shown.  All rankings for all submissions
as well as performance on individual datasets are detailed on the competition and github site.
\vspace{-15pt}} 

{\footnotesize
\begin{tabular}{lrrrrrrr}
    \hline
    \multicolumn{7}{c}{Recall achieved at a minimum of throughput 2K QPS} \\
    \hline
    Algorithm & DEEP &  BIGANN  & MS Turing &  MS SpaceV &  Text-to-image & SSN++ \\
    \hline
    baseline & 0.94275  & 0.93260  & 0.91322 & 0.90853 & 0.86028 & 0.97863 \\
    \hline
    OptaNNE GraphNN & 0.99882 & 0.99978  & 0.99568 &  0.99835 & 0.97340 & - \\
    CUANNS IVFPQ & 0.99543 & 0.99881  & 0.988993 & 0.99429 & 0.94692 & - \\
    CUANNS MultiGPU & 0.99504 & 0.99815 & 0.98399 & 0.98785 & - & - \\
    \hline
  \end{tabular}
}

{\footnotesize
\begin{tabular}{lrrrrrrr}
    \hline
    \multicolumn{7}{c}{Querries per second at 90\% recall} \\
    \hline
    Algorithm & DEEP &  BIGANN  & MS Turing &  MS SpaceV &  Text-to-image & SSN++ \\
    \hline
    baseline & 44464  & 3271 & 2845 & 3265 & 1789 & 5699 \\
    \hline
    CUANNS MultiGPU & 8016944 & 747421 & 584293 & 839749 & - & - \\
    OptaNNE GraphNN & 1965446 & 335991  & 161463 &  157828 & 17063 & - \\
    CUANNS IVFPQ & 91701 & 80109  & 109745 & 108302 & 19094 & - \\
    \hline
  \end{tabular}
}
 
{\footnotesize
\begin{tabular}{lrrrrrrr}
    \hline
    \multicolumn{7}{c}{ Joule/query achieved at minimum of 2K QPS and 0.9 recall} \\
    \hline
    Algorithm & DEEP &  BIGANN  & MS Turing &  MS SpaceV &  Text-to-image & SSN++ \\
    \hline
    baseline & 0.1117  & 0.1576 & 0.1743 & 0.1520 & 0.1128 & 0.0904 \\
    \hline
    OptaNNE GraphNN & 0.00441 & 0.0022  & 0.0048 &  0.0049 & 0.0446 & - \\
    CUANNS IVFPQ & 0.0112 & 0.0112  & 0.0119 & 0.0090 & 0.0480 & - \\
    CUANNS MultiGPU & 0.0029 & 0.0024 & 0.0049 & 0.0023 & - & - \\
    \hline
  \end{tabular}
} 
  
{\footnotesize
\begin{tabular}{lrrrrrrr}
    \hline
    \multicolumn{7}{c}{Total cost to horizontally replicate a system to serve 100\,000 queries per second} \\
    \hline
    Algorithm & DEEP &  BIGANN  & MS Turing &  MS SpaceV &  Text-to-image & SSN++ \\
    \hline
    baseline & 545.6  & 737.9 & 853.9 & 735.9 & 1272.7 & 428.1 \\
    \hline
    OptaNNE GraphNN & 16.1 & 15.4  & 16.3 &  16.4 & 103.6 & - \\
    CUANNS IVFPQ & 303.9 & 304.2  & 153.2 & 153.2 & 916.8 & - \\
    CUANNS MultiGPU & 569.1 & 569.2 & 286.9 & 398.2 & 1213.8 & 629.4 \\
    \hline
  \end{tabular}
}
  
\label{table:t3}
\end{table}

Table~\ref{table:t3} summarizes the results of the competition.
The OptaNNE GraphANN implementation scored the highest on the recall and
cost- and power-normalized leaderboards. The CUANNS MultiGPU implementation
scored the highest on the throughput leaderboard. We did not combine the 
four different benchmark to determine one winner.  We instead chose to name Intel and NVidia 
the co-winners of this track:

\begin{quote}
\textbf{OptaNNE GraphNN}, \emph{Sourabh Dongaonkar (Intel Corporation), Mark Hildebrand (Intel Corporation / UC Davis), Mariano Tepper (Intel Labs), Cecilia Aguerrebere (Intel Labs), Ted Willke (Intel Labs), Jawad Khan (Intel Corporation)}
\end{quote}

\begin{quote}
    \textbf{CUANNS MultiGPU}, \emph{Akira Naruse (NVIDIA), Jingrong Zhang (NVIDIA), Mahesh Doijade (NVIDIA), Yong Wang (NVIDIA), Hui Wang (Xiamen University), Harry Chiang (National Tsing Hua University)}
\end{quote}

\section{Conclusion, Lessons learned, and Outlook}

With this challenge, we started a principled and reproducible research
environment for the design and the evaluation of nearest neighbor implementations. 
In all three tracks we received implementations that were able to beat the well-designed baseline,
which shows that improvements in the state-of-the-art are in reach even in a short time span. 

\paragraph{Lessons learned.} As mentioned by several participants, the design, implementation,
and iterative engineering of nearest neighbor search methods usually takes time that exceeds 
the short time span of such a challenge. For this reason, the evaluation framework, which 
is now well established, will continue to accept new submissions.
Moreover, the evaluation process is time consuming because of the scale of the problem.
Some additional software development is required to further automate the evaluation
and ensure organizers are not the 
bottleneck in the evaluation process.
The cost related to building indices in the cloud is a limit to frequently re-running the 
entire evaluation, and could also be a bottleneck for non-industry affiliated teams
when thinking of going beyond billion-scale datasets.

\paragraph{Future Directions.} Possible future directions were discussed at an open forum
during the NeurIPS break-out session and documented on the evaluation framework
\footnote{\url{https://github.com/harsha-simhadri/big-ann-benchmarks/issues/90}}.
Some questions and directions for new tasks include: 
\begin{itemize}
    \item Support ANNS queries which also allow filters such as date range, author, language, 
    image color or some combination of such attributes. See, for example, \citep{analyticdbv},
    a candidate algorithm for the problem.
    \item Design algorithms whose accuracy and performance is robust to insertions and deletions. A possible strong baseline is Fresh-DiskANN~\citep{FreshDiskaNN}.
    \item Design algorithms that are robust to datasets with out-of-distribution queries such as those arising from cross-modal embeddings.
    \item Design compression with lesser information loss, perhaps at the price of more expensive decoding.
\end{itemize}

\paragraph{Outlook.} We invite researchers and practitioners to use our framework as a starting point,
and look forward to contributions of new datasets and algorithms on an ongoing basis. 
The website \url{https://big-ann-benchmarks.com/} will serve as the main interface, 
and we plan to repeat the challenge in the future, possibly with additional directions.

\section*{Acknowledgements}
We thank all the participants for their submissions to this challenge. 
Moreover, we thank Alexandr Andoni and Anshumali Shrivastava for their
invited talks. We thank Microsoft Research for help in organizing this
competition, and contributing compute credits. We thank Microsoft Bing
and Turing teams for creating datasets for release. Furthermore, we thank
the organizing committee Douwe Kiela, Barbara Caputo, and Marco Ciccone
of the NeurIPS 2021 challenge track for their excellent work.

\bibliography{ref}

\end{document}